%% file: coling_latex.tex
\pdfoutput=1
\documentclass[11pt]{article}

\usepackage[preprint]{coling}

\usepackage{times}
\usepackage{latexsym}
\usepackage{booktabs}
\usepackage{colortbl}
\usepackage{multirow}

\usepackage[T1]{fontenc}
\usepackage[utf8]{inputenc}

\usepackage{microtype}

\usepackage{inconsolata}

\usepackage{graphicx}

\title{MC-CoT: A Modular Collaborative CoT Framework for Zero-shot Medical-VQA with LLM and MLLM Integration}

\author{
  Lai Wei$^{1}$,
  Wenkai Wang$^{1}$,
  Xiaoyu Shen$^{2}$,
  Yu Xie$^{3}$, 
  Zhihao Fan$^{4}$, \\
  \textbf{
  Xiaojin Zhang$^{1}$,
  Zhongyu Wei$^{5}$
  and Wei Chen$^{1}$\thanks{Corresponding author. Email:lemuria\_chen@hust.edu.cn}}
  \vspace{0.5em}  \\   
  $^1$Huazhong University of Science and Technology,\\
  $^2$Eastern Institute of Technology,  $^3$Purple Mountain Laboratories, \\
  $^4$Alibaba Inc., $^5$Fudan University 
}
\begin{document}
\maketitle
\begin{abstract}
In recent advancements, multimodal large language models (MLLMs) have been fine-tuned on specific medical image datasets to address medical visual question answering (Med-VQA) tasks. However, this common approach of task-specific fine-tuning is costly and necessitates separate models for each downstream task, limiting the exploration of zero-shot capabilities. In this paper, we introduce MC-CoT, a modular cross-modal collaboration Chain-of-Thought (CoT) framework designed to enhance the zero-shot performance of MLLMs in Med-VQA by leveraging large language models (LLMs). MC-CoT improves reasoning and information extraction by integrating medical knowledge and task-specific guidance, where LLM provides various complex medical reasoning chains and MLLM provides various observations of medical images based on instructions of the LLM. Our experiments on datasets such as SLAKE, VQA-RAD, and PATH-VQA show that MC-CoT surpasses standalone MLLMs and various multimodality CoT frameworks in recall rate and accuracy. These findings highlight the importance of incorporating background information and detailed guidance in addressing complex zero-shot Med-VQA tasks. Our code is available at \url{https://github.com/thomaswei-cn/MC-CoT}.

% By fine-tuning or pre-training various multimodal large language models (MLLMs) on medical image datasets, significant progress has been made in addressing medical visual question answering (Med-VQA) tasks. 
% However, this pre-training approach is not only costly but also limits the model's generalization capabilities. 
% We believe that focusing on enhancing the model's zero-shot ability for Med-VQA is crucial.
% In this paper, we present the MC-CoT, a modular cross-modal collaboration Chain-of-Thought (CoT) framework designed to enhance MLLM's zero-shot performance on Med-VQA by leveraging large language models (LLMs). 
% MC-CoT enhances the reasoning and information extraction capabilities of MLLMs by integrating medical knowledge and task-specific guidance. 
% Through experiments on datasets such as SLAKE, VQA-RAD, and PATH-VQA, we demonstrate that the MC-CoT framework outperforms both standalone MLLM and other CoT frameworks in terms of recall rate and accuracy.
% Our findings further emphasize the crucial role of providing background information and detailed guidance in tackling highly specialized and complex zero-shot Med-VQA tasks. Our code is available at \url{https://github.com/thomaswei-cn/MC-CoT}.

\end{abstract}

\section{Introduction}

\begin{figure}
    \centering
    \includegraphics[width=1\linewidth]{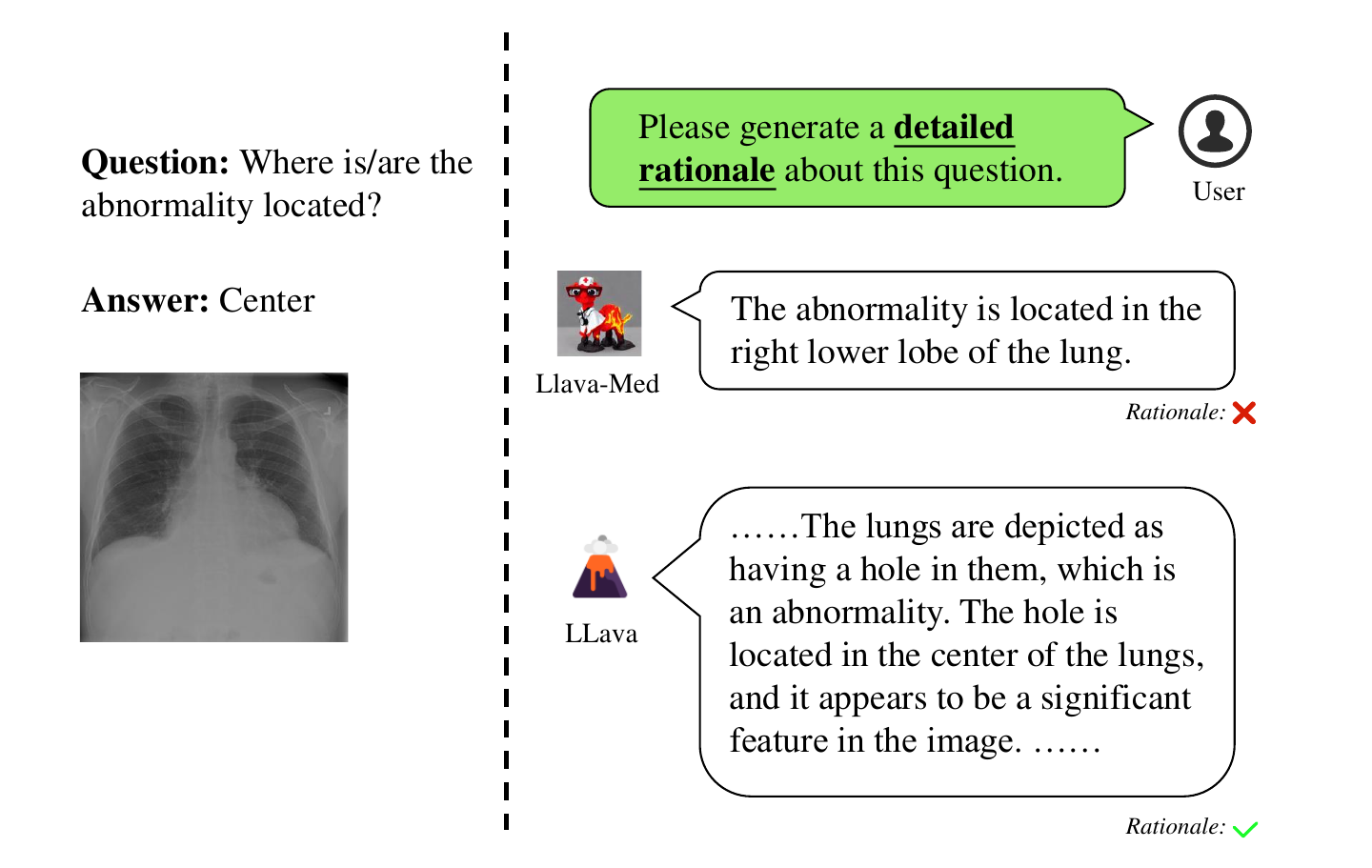}
    % \caption{Current open-source MLLMs in the medical field, whether based on pre-training or fine-tuning, exhibit poor instruction-following capabilities.}
    \caption{The ability of instruction-following for open source multimodal large language models in the medical field is severely degraded.}
    \label{fig:poor_instruct}
\end{figure}

The recent development of large language models (LLMs)~\cite{achiam2023gpt,reid2024gemini,touvron2023llama}, especially multimodal large language models (MLLMs)~\cite{zhu2023minigpt,liu2024visual}, has garnered significant attention in the medical visual question answering (Med-VQA) task~\cite{he2020pathvqa,liu2021slake,zhang2023pmc}. By pre-training on extensive pairs of medical images and text and then fine-tuning on specific Med-VQA datasets, these models~\cite{moor2023med,li2024llava} have achieved promising results. However, this approach adheres to the conventional \emph{pre-training and fine-tuning} paradigm, requiring dedicated models for different task or dataset, thereby consuming substantial resources and limiting scalability, as shown in Figure~\ref{fig:poor_instruct}.

Med-VQA is a complex and challenging multimodal task~\cite{lin2023medical} that necessitates extensive medical knowledge in areas such as anatomy, pathology, and clinical medicine to accurately interpret the rich and specialized content of medical images, including X-rays, CT scans, and MRI scans. Additionally, this task requires sophisticated multimodal reasoning abilities~\cite{zhang2023multimodal} to integrate and analyze data spanning visual, textual, and other formats. These abilities encompass advanced modality fusion, contextual understanding, and language processing, all crucial for delivering precise and clinically relevant answers. However, the capability of LLMs or MLLMS to perform Med-VQA tasks without task-specific fine-tuning or zero-shot capabilities, along with their proficiency in multimodal reasoning on medical images, remains largely unexplored.

% Inspired by research on LLM-powered multi-agent system in the medical field~\cite{tang2023medagents,fan2024ai}, in this paper, 

% Inspired by research on multi-agent systems in the medical field driven by LLMs in a purely textual modality~\cite{tang2023medagents,fan2024ai}, this paper conducts an extensive study on a multi-modal reasoning framework for medical imaging.

% we propose \textbf{MC-CoT}, a framework to enhance the zero-shot performance on Medical VQA tasks.

Inspired by recent research on intelligent systems in the medical field driven by LLMs in purely textual modalities~\cite{bao2023disc,tang2023medagents,fan2024ai}, this paper presents an extensive study on a multi-modality reasoning framework tailored for medical imaging question answering. We introduce \textbf{MC-CoT}, a \textbf{M}odular \textbf{C}ollaborative \textbf{C}hain-\textbf{o}f-\textbf{T}hought framework designed to enhance the zero-shot performance of MLLMs on Med-VQA tasks. MC-CoT integrates LLMs into the problem-solving process, utilizing their extensive knowledge and robust chain-of-thought (CoT) reasoning capabilities to guide the analysis and response generation. Specifically, MC-CoT comprises three pre-designed image feature extraction modules—\emph{Pathology}, \emph{Radiology}, and \emph{Anatomy}. Each module targets specific aspects of the images, designed to process particular tasks related to its focus area and generate informed responses based on the image data. Within these modules, the LLM first evaluates the input task, supplying essential background knowledge and strategic guidance to the MLLM, which then produces the final output. When faced with a new problem, MC-CoT leverages the LLM to dissect the issue, activating one or more of its specialized modules based on the problem's requirements. Each module is assigned specific tasks, after which the LLM synthesizes the outputs from the engaged modules to formulate a cohesive and comprehensive final answer.

We evaluate MC-CoT's performance on three diverse Med-VQA datasets: PATH-VQA, VQA-RAD, and SLAKE, comparing it against baseline visual CoT methods and other collaborative frameworks. Our experiments demonstrate that MC-CoT consistently outperforms existing approaches in terms of answer accuracy and recall of key information. The framework's effectiveness is verified across different MLLM and LLM combinations, highlighting its broad applicability. Further analysis reveals the significant impact of key processes within MC-CoT, such as image captioning, LLM-guided reasoning, and answer summarization. These findings suggest that MC-CoT offers a promising approach for enhancing the zero-shot capabilities of MLLMs in Med-VQA tasks by effectively integrating domain-specific knowledge and guided reasoning.

In summary, our paper makes the following two key contributions:

\begin{itemize}
    \item We propose MC-CoT, a novel framework that enhances the zero-shot performance of Multimodal Large Language Models (MLLMs) on Med-VQA tasks. 
    \item We conduct comprehensive experiments on diverse Med-VQA datasets, our extensive analysis demonstrates MC-CoT's broad applicability and provides insights into the impact of key processes and specialized modules, offering directions for future optimization in Med-VQA systems.
\end{itemize}

\begin{figure*}[t]
    \centering
    \includegraphics[width=1\linewidth]{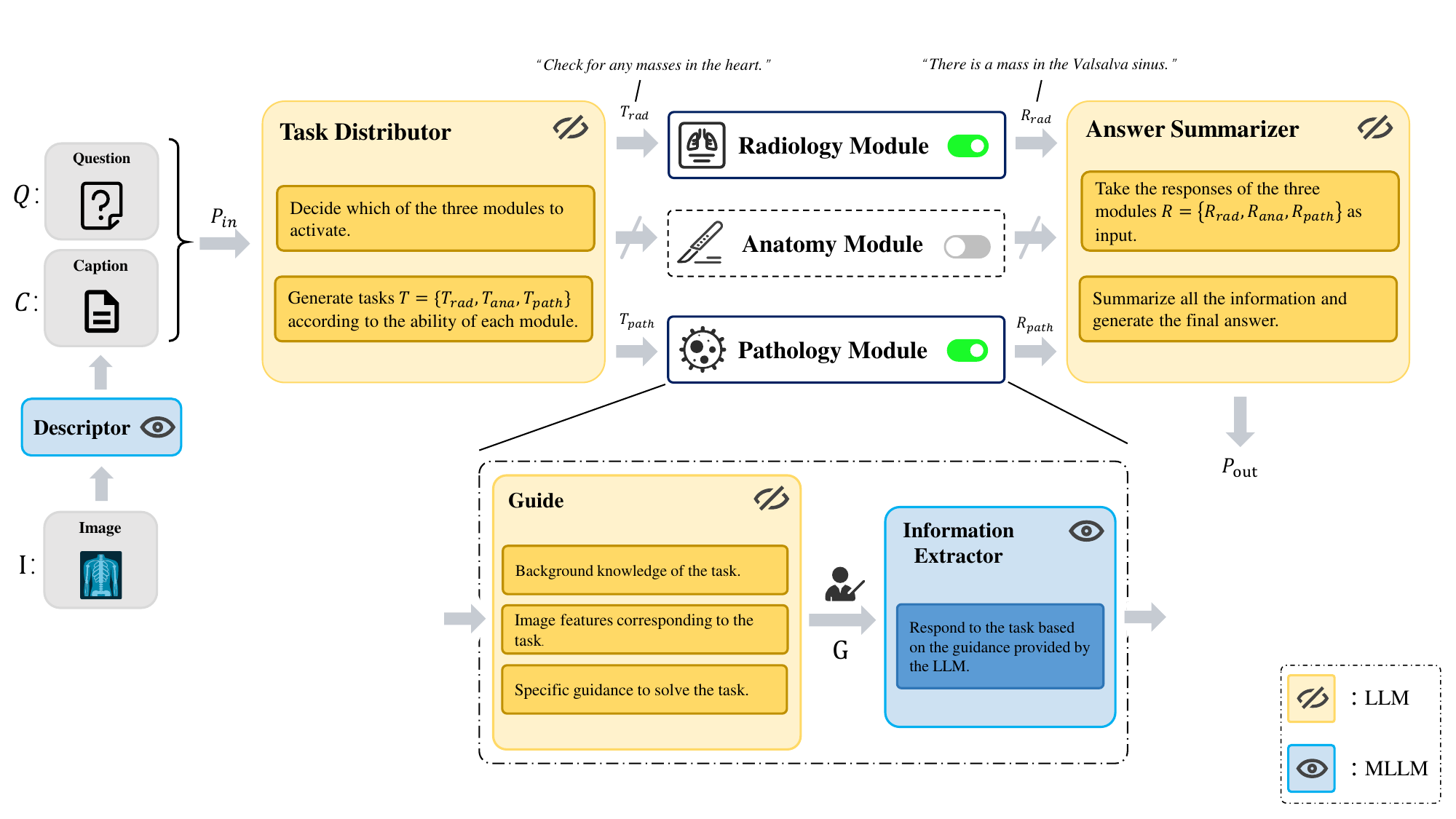}
    \caption{The schematic diagram of the MC-CoT framework.}
    \label{fig:M3}
\end{figure*}

\input{related}

\input{method}

\input{experiment_setup}

\input{main_results}

\section{Conclusion}

The paper explores Med-VQA tasks with open-ended questions by experimenting with various CoT frameworks, investigating the feasibility of enhancing MLLM's zero-shot capabilities without the need for pretraining or fine-tuning. We introduce MC-CoT, a modular approach that examines problems from multiple angles and uses LLM to supply the necessary background knowledge and solutions for MLLM to address the tasks effectively. MC-CoT outperforms both directly use MLLM and several popular CoT frameworks, including MMCoT and DDCoT, in terms of recall rate and accuracy. Furthermore, MC-CoT exhibits strong generalizability, maintaining its effectiveness across various LLM and MLLM combinations.

\section*{Limitation}

While the MC-CoT framework demonstrates significant improvements in enhancing MLLM’s zero-shot performance on Med-VQA tasks, it is not without limitations.

Firstly, the three modules we designed—\emph{anatomy}, \emph{pathology}, and \emph{radiology}—have only been validated on open-ended questions in the PATH-VQA, VQA-RAD, and SLAKE datasets. 
When encountering new problems, a substantial redesign of the modules and their associated prompts may be required.

Moreover, the effectiveness of the MC-CoT framework heavily relies on the quality of the prompts provided by the LLM to guide the MLLM. 
Crafting prompts that effectively guide the model without overly restricting its reasoning can be challenging and may require significant domain expertise. 
Additionally, the LLM itself must possess sufficient medical knowledge.

Currently, the framework only considers image-based information, without incorporating other clinical symptoms such as patient history, which is crucial for real-world applications. 
Integrating more comprehensive patient data could enhance the framework’s performance but would also introduce additional complexity.

Lastly, our evaluation metrics are tailored to Med-VQA and may not fully capture the nuances of real-world medical diagnostic reasoning. 
As application scenarios change, further research will be necessary to develop evaluation methods that better reflect clinical utility.

\bibliography{custom}
\appendix
\input{appendix}

\end{document}

%% file: related.tex
\section{Related Works}

\subsection{Medical Visual Question Answering}

% The task of Med-VQA is to provide answers to questions posed about medical images. 
% Developing models or methods capable of answering questions based on biomedical images can offer support to clinicians and patients. 
% Earlier work predominantly treated Med-VQA as a simple classification problem\citep{eslami2021does,li2023self,gong2022vqamix,do2021multiple,zhang2023biomedclip}, achieving high accuracy on small-scale datasets. 
% However, this approach struggles with the diverse expressions of questions it may encounter in real-world scenarios. 

% Recently, using image-text pairs to train MLLM models and fine-tune them for downstream tasks has become the preferred approach for most multimodal tasks.
% Researchers have begun to adopt generative approaches to tackle Med-VQA\citep{chen2024miss,van2023open}. 
Prior to the development of Med-VQA, research in medical question answering focused on text-based datasets~\cite{jin2019pubmedqa,pal2022medmcqa,chen2023benchmark,chen2023dxformer,chen2023knse} which facilitated advancements in medical natural language processing. The Med-VQA task involves answering questions about medical images, which can provide clinicians with more fine-grained references. Early work primarily viewed Med-VQA as a classification problem~\citep{eslami2021does,li2023self,gong2022vqamix,do2021multiple,zhang2023biomedclip}, achieving high accuracy on small datasets, but struggled with the diverse expressions of real-world questions. 

Recently, training MLLMs with image-text pairs for downstream tasks has become the preferred approach, leading researchers to adopt generative methods for Med-VQA~\citep{chen2024miss,van2023open}.

\subsection{Multimodal Prompting Methods}

As LLMs and MLLMs gain popularity, effective prompting techniques have become essential to fully leverage their capabilities. These approaches generally fall into three categories: zero-shot prompting~\citep{kojima2022large,wan2023better}, few-shot prompting~\citep{brown2020language,dong2022survey,min2022rethinking}, and other methods like ToT~\citep{yao2024tree} and GoT~\citep{besta2024graph,zheng2024reverse}.

In multimodal domains, various prompting techniques have emerged. MM-CoT~\citep{zhang2023multimodal}, CCoT~\citep{mitra2024compositional}, and the Cantor framework~\citep{gao2024cantor} rely exclusively on MLLMs. MM-CoT uses a two-stage approach: generating rationales based on ground truth before producing the final answer. Similarly, CCoT generates a scene graph to aid reasoning. Cantor constructs modular structures for problem decomposition. 

DDCoT~\citep{zheng2023ddcot}, on the other hand, uses LLM to assist MLLM by breaking the original question into sub-questions to form a reasoning process.

\subsection{Multi-modal Large Language Models}

Multimodal Large Language Models~\citep{chowdhery2023palm,li2022blip,radford2021learning} capitalize on the cognitive abilities of LLMs to enhance reasoning in tasks that require both visual and textual understanding, such as Visual Question Answering (VQA)~\citep{antol2015vqa,hudson2019gqa,marino2019ok}. 
These models typically bridge pre-trained visual representations with pre-trained LLMs through additional layers, enabling them to exhibit visual perception abilities comparable to those of humans, along with some degree of logical reasoning.

Despite their potential, the number of MLLMs specifically tailored for the medical domain remains limited. 
Among the available open-source models, LLaVA-Med and Med-Flamingo~\citep{moor2023med} stand out, but they are constrained by their relatively small parameter sizes. 
This limitation results in significant shortcomings in their logical reasoning and instruction-following capabilities, especially when compared to other open-source MLLMs.

Nevertheless, while recent open-source MLLMs such as LLaVA-1.5~\citep{liu2024improved} and DeepSeek-VL~\citep{lu2024deepseek} have made substantial strides in improving instruction execution and image perception, they still face considerable challenges in the medical domain due to the lack of domain-specific knowledge. 

%% file: method.tex
\section{Method}
\input{Tables/symbols}

% Overall, the collaboration methods between the MLLM and LLM are very flexible and diverse. 
% There are many ways to cooperate MLLM with LLM.
% Relevant studies, such as DDCoT and IdealGPT \citep{you2023idealgpt}, have already demonstrated that in general domain VQA tasks, using LLM to break down the problem is a feasible way to stimulate the model's latent reasoning capabilities. 
There are various ways to cooperate MLLM with LLM. 
Studies like DDCoT and IdealGPT\citep{you2023idealgpt} have shown that leveraging LLM to decompose tasks effectively enhances reasoning in general domain VQA.
Based on this idea, we equipped the MC-CoT framework with three specialized modules, each with clearly defined functions, to guide the LLM in decomposing problems along the aspects that each module focuses on.

Additionally, although general-domain MLLMs have demonstrated impressive reasoning abilities on tasks such as ScienceQA \citep{saikh2022scienceqa,schwenk2022okvqa}, they struggle to interpret medical images like doctors due to their limited medical knowledge and weak medical reasoning abilities. 
Therefore, we consider leveraging LLMs to provide guidance, compensating for the deficiencies of MLLMs in reasoning abilities.

In general, as shown in Figure \ref{fig:M3}, our MC-CoT framework is divided into three stages: \textit{Module Activation and Task Assignment}, \textit{Modular Medical Image Feature Extraction}, and \textit{Answer Generation}. The symbols and their meanings used in this paper to illustrate the frameworks are shown in the table \ref{tab:symbol}. The specific prompts are illustrated in Appendix~\ref{appendix:prompts}.

\subsection{Module Activation and Task Assignment} 

Inspired by~\cite{chen2023disc}, the goal of this stage is to determine the problem-solving approach based on the problem scenario and assign specific tasks to each module. Given the LLM's strong reasoning abilities and extensive background knowledge, we chose to use the LLM to complete the aforementioned process. 
However, since the LLM cannot directly interpret images, providing only the question ($Q$) might lead the LLM to make random guesses about the image content based on the question's context, which could negatively impact the subsequent reasoning process. Therefore, we also include an image caption ($C$) as input to further narrow down the scope of the question.

Specifically, we use the prompt \emph{Please provide a detailed description of the features you believe are relevant to the question} to instruct the MLLM to generate a simple and objective description of the image, referred to as $C$. The MC-CoT then combines $C$ with the original question $Q$ to form $P_{in} = \{Q, C\}$, which is then input into the LLM. 
In the LLM, we applied the CoT method to improve the quality of its output. 
It first generates a rationale, then selects and activates the appropriate modules, and finally generates tasks $T=\{T_{rad},T_{ana},T_{path}\}$ for each activated module.

\subsection{Modular Medical Image Feature Extraction}

This stage is equipped with three modules, each with distinct capabilities and responsibilities:

\textbf{Radiology Module:}\quad Determine the appropriate imaging modality, identify the imaging plane, locate the lesion, and analyze the color/contrast characteristics to differentiate tissue types and abnormalities.

\textbf{Anatomy Module:}\quad Identify the organ or anatomical structure involved and provide detailed information on the anatomical position and relations of the lesion within the body. 

\textbf{Pathology Module:}\quad Consider the number of lesions and their clinical significance, and provide a reasonable explanation for the phenomenon using pathology knowledge.

Based on the tasks $T=\{T_{rad},T_{ana},T_{path}\}$ provided by the Module Activation and Task Assignment module, MC-CoT utilizes these three specialized modules to extract and analyze information from the image. 
Since we are discussing the zero-shot capabilities of MLLM, we use a general-domain MLLM to extract additional information from the image.

To tackle the lack of medical background in general-domain MLLM, each module incorporates an LLM-based process. Using the task from the preceding module, we input the following instructions into the LLM:
\begin{itemize}
    \item \textit{Please use your medical knowledge to provide a guide on how to solve the task.}
    \item \textit{You need to explain the features that the image may contain based on the task, and how to give the right answer from the perspective of the picture.}
\end{itemize}
These instructions prompt the LLM to analyze the task and provide a detailed guide $G$. Additionally, we remind the LLM:
\begin{itemize}
    \item \textit{Remember you are teaching a rookie to read a medical image. So make sure you break down medical or biological terms into intuitive descriptions, especially terms related to image features.}
\end{itemize}
This reminder ensures that the LLM's response minimizes the use of complex medical terminology and makes the explanation more accessible.
\begin{itemize}
    \item \textit{You cannot give your speculation on the final answer.}
\end{itemize}
This reminder instructs the LLM to remain neutral and avoid making assumptions about the final answer.

This process above provides guidance and relevant background information to the MLLM, enabling it to focus on specific features of the image to better complete the task. Subsequently, the MLLM takes $\{I, T_x, G\}$ as input, extracts information from the image based on the guidance, and outputs the final answer.

\subsection{Answer Generation:} 
The last module is responsible for logical reasoning based on the question and synthesizing all supplementary information extracted from the image $R=\{R_{path}, R_{ana}, R_{rad}\}$ by three specialized modules , along with the image's caption $C$, to ultimately provide an answer.

MC-CoT uses an LLM to complete this process. 
To further enhance the logical reasoning ability of LLM, we require LLM to generate a reasoning process before outputting the final answer.

%% file: Tables/symbols.tex
\begin{table*}
\centering
\begin{tabular}{c|l} 
\toprule
\textbf{Symbol}& \textbf{Meaning}\\
\hline
$Q$ & Question \\
$C$ & Caption of the image \\
$I$ & Image \\
$G$ & Guide generated by LLM \\
$P_{in},P_{out}$ & Input and output of the whole framework \\
$T_{rad},T_{ana},T_{path}$ & Tasks assigned to the radiology, anatomy and pathology module \\
$R_{rad},R_{ana},R_{path}$ & Responses to the task from the radiology, anatomy and pathology module  \\
\bottomrule
\end{tabular}
\caption{Explanation of the meanings behind the symbols used.}
\label{tab:symbol}
\end{table*}

%% file: experiment_setup.tex
\section{Experiment Setup}

\subsection{Datasets}
We evaluated MC-CoT's performance on three Med-VQA datasets. All three datasets include open-ended questions (e.g., why, what, how, where) and closed-ended questions (mostly yes/no). The ground truth for open-ended questions primarily consists of single nouns or phrases, such as "Chest" or "Lung Cancer."

Basic information about the datasets is summarized in Table~\ref{tab:dataset info}.

\begin{itemize}
    \item \textbf{PATH-VQA}\quad This is a pathology image dataset. 
    It includes 4,998 images paired with 32,799 QA pairs. Each image is linked to questions about location, shape, color, and appearance. 
    Answers require not just image analysis but also integration with pathological knowledge to derive conclusions.
    \item \textbf{VQA-RAD}\quad It features 3,515 clinician-created QA pairs and 315 radiology images, evenly distributed among the head, chest, and abdomen. 
    The questions cover 11 categories: abnormalities, modalities, organ systems, color, count, object/condition presence, size, plane location reasoning, and others.
    \item \textbf{SLAKE}\quad It contains 642 radiology images and over 7,000 QA pairs, featuring diverse modalities across body parts like the brain, neck, chest, abdomen, and pelvis. 
    Additionally, SLAKE is bilingual with English and Chinese, but our experiments used only the English data.
\end{itemize}

The ultimate goal of developing a Med-VQA system is to answer user questions about a medical image in natural language. 
To align with this goal, we conducted experiments using only open-ended questions from the test sets of the datasets mentioned.

\input{Tables/dataset_info}

\subsection{MC-COT Settings}

% For the selection of MLLMs, LLaVA-1.5-7B was used for the primary experiments and analysis. 
% In the experiments, the precision of LLaVA-1.5-7B was set to fp16. 
% To verify the broad applicability of our MC-CoT framework, we also conducted supplementary experiments using DeepSeek-VL-7B-chat \citep{lu2024deepseek}, Qwen-VL-Chat\citep{bai2023qwen} and Qwen-VL-Max. 

% As for the selection of LLMs, we used GPT-3.5 to play the role of LLM in the MC-CoT framework for the main experiments. 
% In supplementary experiments, we conducted tests using GLM-4-9B-Chat\citep{glm2024chatglm}, Qwen2-72B-Instruct\citep{qwen2}, and Deepseek-V2\citep{deepseekai2024deepseekv2strongeconomicalefficient}, respectively.
For MLLM selection, LLaVA-1.5-7B was used in the primary experiments and analysis, with its precision set to fp16. To verify the broad applicability of our MC-CoT framework, we also conducted supplementary experiments using DeepSeek-VL-7B-chat\citep{lu2024deepseek}, Qwen-VL-Chat\citep{bai2023qwen}, and Qwen-VL-Max. 

For LLM selection, GPT-3.5 was used in the main experiments within the MC-CoT framework. 
In supplementary experiments, we tested GLM-4-9B-Chat\citep{glm2024chatglm}, Qwen2-72B-Instruct\citep{qwen2}, and Deepseek-V2\citep{deepseekai2024deepseekv2strongeconomicalefficient}, respectively.

\input{Tables/main_results}

\subsection{Model Comparison}

To further explore the capabilities of the MC-CoT framework, we compared it with other frameworks, including two simple and intuitive MLLM-LLM collaborative CoT frameworks we designed: IICoT and FCCoT.

\textbf{IICoT}\quad The name IICoT stands for \textbf{I}nformation-\textbf{I}nstruction \textbf{C}hain-\textbf{o}f-\textbf{T}hought, which refers to using an LLM to supplement the current problem context with background information and provide guidance for solving the problem. 
The MLLM then follows this guidance to answer the question. 

\textbf{FCCOT}\quad The name FCCoT stands for \textbf{F}law-\textbf{C}heck \textbf{C}hain-\textbf{o}f-\textbf{T}hought, which refers to first using an MLLM to generate an analysis of the problem, then using both the LLM and MLLM to check for flaws in the reasoning, factual information, and other aspects of the analysis.
Finally, the MLLM revises its analysis and re-generates the answer.

Details about these two frameworks can be found in the appendix~\ref{appendix:other_frameworks}.

For the \emph{Question-Driven Image Captions as Prompts} \citep{ozdemir2024enhancing} method, which we refer to as QD Cap., we use the MLLM to generate image captions and the LLM as the QA model to provide answers.

Furthermore, for a fairer comparison, we replaced the four general-purpose expert modules in Cantor with the three medical expert modules from the MC-CoT.

\subsection{Evaluation Metrics}
We proposed two evaluation methods: one is an automated evaluation metric, and the other is a model-based metric.

\textbf{Automated Evaluation}\quad Since the ground truth answers in our dataset are single nouns or phrases, we believe that the presence of key words from the correct answers in model-generated responses indicates that the model has focused on essential information. 

Thus, we calculate the \emph{recall} rate of correct answers within the generated responses to assess the model's ability to concentrate on key information in the question context.

\textbf{Model-based Evaluation}\quad Considering that the same medical concept can be expressed in different ways, a simple string comparison is insufficient. 
Instead, we use an LLM to assess the conceptual overlap between the generated answers and the ground truth to evaluate the \emph{accuracy}. 

We utilize Deepseek-V2 to rate the model's responses on a scale of 1 to 4, with 1 being the worst and 4 the best, and the final scores are scaled to a maximum of 100 points.

The implementation details of the evaluation and the prompts we used can be found in Appendix~\ref{appendix:prompts_eval}.

%% file: Tables/dataset_info.tex
\begin{table}[h]
\centering
\begin{tabular}{lrrr}
\toprule
Dataset  & Images & QA pairs & \# \\ 
\midrule
VQA-RAD & 0.3k  & 3.5k &  949    \\
PATH-VQA  & 5k & 32.8k  & 625   \\
SLAKE & 0.7k  & 14k & 645  \\                     
\bottomrule
\end{tabular}
\caption{Basic information about the dataset used for evaluation. \# represents the number of open-ended QA pairs in the test sets used in our experiments.}
\label{tab:dataset info}
\end{table}

%% file: Tables/main_results.tex
\begin{table*}
\centering
\begin{tabular}{l|cccccc|cc}
\toprule
\multicolumn{1}{l|}{}& \multicolumn{2}{c}{PATH-VQA} & \multicolumn{2}{c}{VQA-RAD} &\multicolumn{2}{c|}{SLAKE} &\multicolumn{2}{c}{Avg.}\\
& Recall & Acc.& Recall & Acc.& Recall & Acc.& Recall&Acc.\\
\midrule
\multicolumn{9}{c}{MLLM alone}\\   
\midrule
Only$^*$ & 44.89 & 26.19  & 51.31 & 32.53 & \textbf{69.96} & 52.09 &55.39 &36.94 \\ 
PS-Prompting\citeyearpar{wang2023plan} & 42.86& 10.35& 48.19& 12.82  &60.19 & 23.62 &50.41 &15.60\\ 
Cantor-med$^{**}$ & 43.41& 26.83  & 49.96& 25.29  & 61.74& 26.20 &51.70 &26.11\\ 
CCoT\citeyearpar{mitra2024compositional}& 47.76& 26.93 & 53.41& 32.35 &67.35 & 47.96 &56.17 &35.75 \\ 
Visual CoT\citeyearpar{shao2024visual}& 44.92& 27.20 & 51.13& 31.05 &66.05 & 47.08 &54.03&35.11  \\ 
MMCoT\citeyearpar{zhang2023multimodal}& 47.64& 27.89 & 52.93& 31.51 &68.30 & 51.73 &56.29 &37.04\\ 
\midrule
\multicolumn{9}{c}{LLM \& MLLM}\\ 
\midrule
DDCoT\citeyearpar{zheng2023ddcot}& 38.96& 37.17 &47.95 &35.09  & 66.07& 47.55 &50.99&39.94\\ 
QD Cap.\citeyearpar{ozdemir2024enhancing}&37.45 & 21.39 & 45.40& 25.78 & 62.55& 45.37 &48.47&30.85  \\ 
IdealGPT\citeyearpar{you2023idealgpt}&42.19 & 19.04 & 50.48&27.50 &66.06 & 44.75 &52.91 &30.43 \\ 
QVix\citeyearpar{yang2023good}&49.59 & 34.88 &53.31 & 32.56 &68.05 &48.11 &56.98 &38.52\\  \midrule
FCCoT & 46.07& 26.83 & 51.91& 27.08 & 65.04&38.76&54.34 &30.89 \\ 
IICoT & 48.92& 36.43 &54.02 &35.65 & 67.68&46.15 &56.87 &39.41 \\ 
\textbf{MC-CoT\ (Ours)}& \textbf{49.90} & \textbf{45.07} & \textbf{57.06} &\textbf{38.25} &69.82 & \textbf{54.88} &\textbf{58.93} &\textbf{46.07} \\
\bottomrule
\end{tabular}
\caption{Using LLaVA-1.5-7B as the MLLM and GPT-3.5 as the LLM, we compared the performance of multiple frameworks across 3 datasets. 
In this table, Only$^*$ means using LLaVA-1.5-7B only.
Cantor-med$^{**}$ is a framework obtained by replacing the four original modules in Cantor with three modules from MC-CoT.}
\label{tab:llava-cp-only}
\end{table*}

%% file: main_results.tex
\input{Tables/change_process}
\section{Experiment Results}

\subsection{Comparison with other CoT frameworks}

\textbf{Improvements in Correctness}\quad The experimental results in Table \ref{tab:llava-cp-only} demonstrate that the MC-CoT framework surpasses most baseline methods in terms of the \emph{recall} rate of correct answers. 
This indicates that the MC-CoT framework is more effective in capturing the key information present in the question’s context, allowing the model to provide more accurate responses. 

In terms of \emph{accuracy}, the MC-CoT framework consistently outperforms LLaVA-1.5-7B and all other CoT frameworks evaluated in this study. 
The results indicate that MC-CoT has enhanced the model's zero-shot capability on the Med-VQA task. 
It also demonstrates that providing necessary background knowledge and guidance to MLLMs can improve their performance to a certain extent, which offers a direction for future research.

\textbf{Effectiveness Across Different LLM and MLLM Combinations}\quad We experimented with different LLM and MLLM pairings to explore the broad applicability of MC-CoT. 
Table \ref{tab:change-mllm} shows the results of supplementary experiments conducted using various MLLMs paired with GPT-3.5. 
The results demonstrate that regardless of the underlying architecture or parameter size of the MLLM, the MC-CoT framework consistently improves the correctness of generated answers. 

Table \ref{tab:change-llm} presents the results of pairing different LLMs with LLaVA-1.5-7B. 
Similarly, whether compared to directly using LLaVA from Table \ref{tab:llava-cp-only} or other LLM-MLLM collaborative frameworks in the experiments, MC-CoT consistently achieved the highest average accuracy.

This consistency across different models indicates that M3 is highly adaptable and capable of enhancing performance across various model combinations.

Importantly, the generalizability of the MC-CoT framework shows that its design principles can be effectively applied to models of different sizes and architectures. This means that MC-CoT does not overly depend on specific model configurations, making it versatile and scalable for Med-VQA.

\subsection{Impact of Key Processes}
Table \ref{tab:change-process} clearly demonstrates the impact of key steps at each stage of the MC-CoT framework. 

\textbf{Captions Help the LLM to Grasp the Context}\quad
The comparison results show that using the image caption along with the question as input achieved higher accuracy than using the question alone. 
This indicates that effectively conveying visual information can enhance the LLM's understanding and reasoning abilities.

\textbf{LLM-Guided Approach Enhances Accuracy, Especially on Challenging Datasets}\quad
Compared to directly using the MLLM to complete the assigned task, the LLM-guided approach improved the overall accuracy. 

Additionally, by observing the degree of accuracy decline, we found that providing guidance has a more significant impact on more challenging datasets such as PATH-VQA and VQA-RAD.

\textbf{LLM Summarization Boosts Performance}\quad
Due to limitations in input length and model capability, the LLM was better able to summarize the collected information and provide answers through reasoning. 
In our experimental results, using the LLM to summarize answers indeed improved the overall efficiency of the MC-CoT framework.

This division of roles between LLM and MLLM optimized their respective strengths, resulting in a better solution for handling complex open-ended Med-VQA problems.

\subsection{Impact of Modules}
\begin{figure}[h!]
    \centering
    \includegraphics[width=1\linewidth]{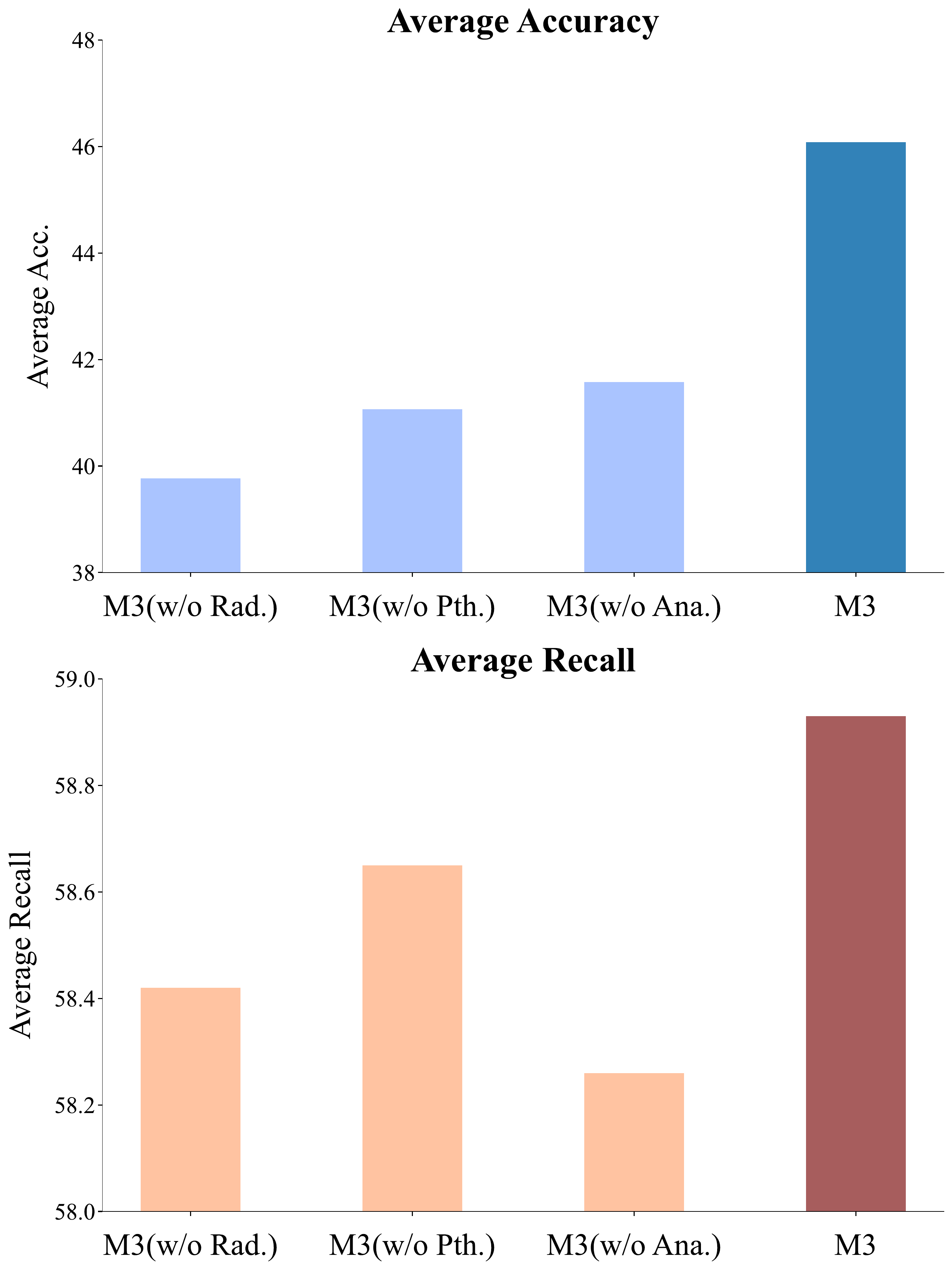}
    \caption{The impact of the 3 modules on the effectiveness of the MC-CoT framework.}
    \label{fig:change-module}
\end{figure}
As shown in Figure~\ref{fig:change-module}, the complete MC-CoT framework achieves the highest average \emph{accuracy} and average \emph{recall} across the three datasets. 

The extent of accuracy decline suggests that the radiology module within the MC-CoT framework contributes the most to the \emph{accuracy} of the answers. 
This implies that determining the imaging modality in advance is crucial when analyzing medical images, as different imaging modalities may highlight different features of the target.

In terms of average \emph{recall}, the anatomy module is the most important. 
This indicates that identifying which body part or organ system is relevant to the question is critical for capturing key information.

\subsection{Case study}
The detailed visual comparison of the output obtained using the MC-CoT framework, the output from other CoT frameworks, and the output from using MLLM alone can be found in Appendix~\ref{appendix:cases}.

Figure \ref{fig:case_path} presents a case from the PATH-VQA dataset. 
Due to the unique nature of the PATH-VQA dataset, where the answers cannot be directly derived from the images but require reasoning based on pathological knowledge, models without access to background knowledge can only provide a vague answer, such as \emph{positively charged protein} (highlighted in yellow). 
In contrast, with DDCoT and MC-CoT, the LLM considered the underlying process behind the question and either proposed guiding questions or directly provided guidance, allowing the answer to be refined to \emph{histone proteins} (highlighted in blue).

Figure \ref{fig:case_slake} highlights the importance of captions. 
While the LLM Guide approach offers instructions for distinguishing different MRI image weights, it lacks targeted guidance due to the LLM's lack of prior knowledge of the image. In contrast, the MC-CoT framework generates a caption for the image initially, which narrows the problem scope and enhances guidance relevance.

In this case, LLaVA-1.5-7B initially lacked knowledge about MRI weights but successfully identified the liver in the image, allowing GPT-3.5 to generate more specific guidance. 
Eventually, LLaVA-1.5-7B learned from the instructions and provided the correct answer.

Finally, in the example shown in Figure \ref{fig:case_rad}, most CoT frameworks fail to comprehend the question (highlighted in yellow). 
The DDCoT framework, however, assumes that sub-question 1 does not require any reliance on visual information and directly provides an answer based on commonsense reasoning from the LLM. 
However, for serious medical issues, we do not want the model to draw conclusions based on common sense and experience. 

Among these methods, only the MC-CoT framework points out the characteristics of the hemidiaphragm as well as the liver and guides the MLLM to search for the corresponding structures in the image for verification (highlighted in blue). 
Clearly, the reasoning process of the MC-CoT framework is more rigorous.

%% file: Tables/change_process.tex
\begin{table*}
\centering
\begin{tabular}{l|cccccc|cc}
\toprule
\multicolumn{1}{l|}{}& \multicolumn{2}{c}{PATH-VQA} & \multicolumn{2}{c}{VQA-RAD} &\multicolumn{2}{c|}{SLAKE} &\multicolumn{2}{c}{Avg.}\\
& Recall & Acc.& Recall & Acc.& Recall & Acc.& Recall&Acc.\\
\midrule
MC-CoT\textit{(w/o caption)} &48.71 & 39.95 &56.02 & 33.02 & 68.23& 48.84 &57.65 &40.60 \\
MC-CoT\textit{(w/o guide)}&49.11 & 32.64 &55.65 & 31.30 & 69.93& 52.61 &58.23 &38.85 \\
MC-CoT\textit{(MLLM summarize)}& 49.82& 28.21 &56.37 & 29.40 &\textbf{69.99} & 49.46 &58.73 &35.69  \\
MC-CoT& \textbf{49.90} & \textbf{45.07}&\textbf{57.06} & \textbf{38.25} &69.82 & \textbf{54.88} &\textbf{58.93} &\textbf{46.07} \\
\bottomrule
\end{tabular}
\caption{The impact of key processes on the effectiveness of the MC-CoT framework.}
\label{tab:change-process}
\end{table*}

%% file: appendix.tex
\section{Supplementary Experiments}
\label{appendix:supplement_experiments}
In the supplementary experiments, we used different combinations of LLMs and MLLMs. 

Table~\ref{tab:change-mllm} presents the results of experiments where we paired two 7B MLLMs (Qwen-VL-Chat and Deepseek-VL-7B) and one MLLM with a larger scale of parameter (Qwen-VL-Max) with GPT-3.5. 

Table~\ref{tab:change-llm} shows experiment results using the GLM-4-9B-Chat, the Qwen2-72B-Instruct, and the Deepseek-V2 model, which has capabilities comparable to GPT-4, all paired with LLaVA-1.5-7B.
\input{Tables/change_mllm}
\input{Tables/change_llm}

\section{Frameworks Designed for Comparison}
\label{appendix:other_frameworks}
\begin{figure*}
    \centering
    \includegraphics[width=1\linewidth]{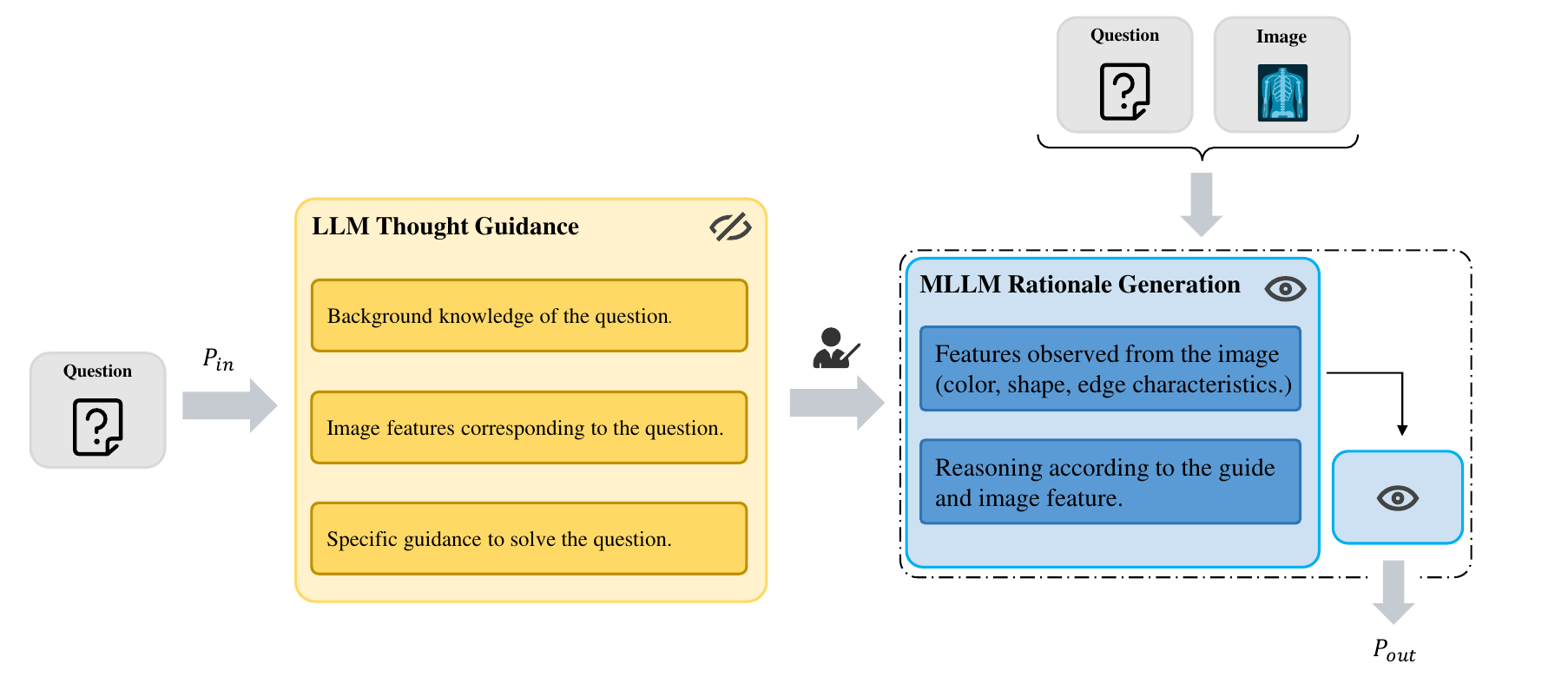}
    \caption{The schematic diagram of the IICoT framework.}
    \label{fig:llm-guide}
\end{figure*}

\textbf{IICoT}\quad This framework consists of one LLM process and one MLLM process, as depicted in Figure~\ref{fig:llm-guide}.
In the first stage, the LLM only takes the text-based question as input. 
We use the same prompt as in the \textbf{Modular Medical Image Feature Extraction} step of the MC-CoT to have the LLM provide supplementary information and guidance to the MLLM. 
The MLLM then follows the LLM's instructions to complete the task.
Specifically, the MLLM step here adopts the same approach as MMCoT to generate a more accurate final answer. 
This involves first using the MLLM to output a rationale, and then re-inputting this rationale into a new MLLM step to generate the final answer.

\begin{figure*}
    \centering
    \includegraphics[width=1\linewidth]{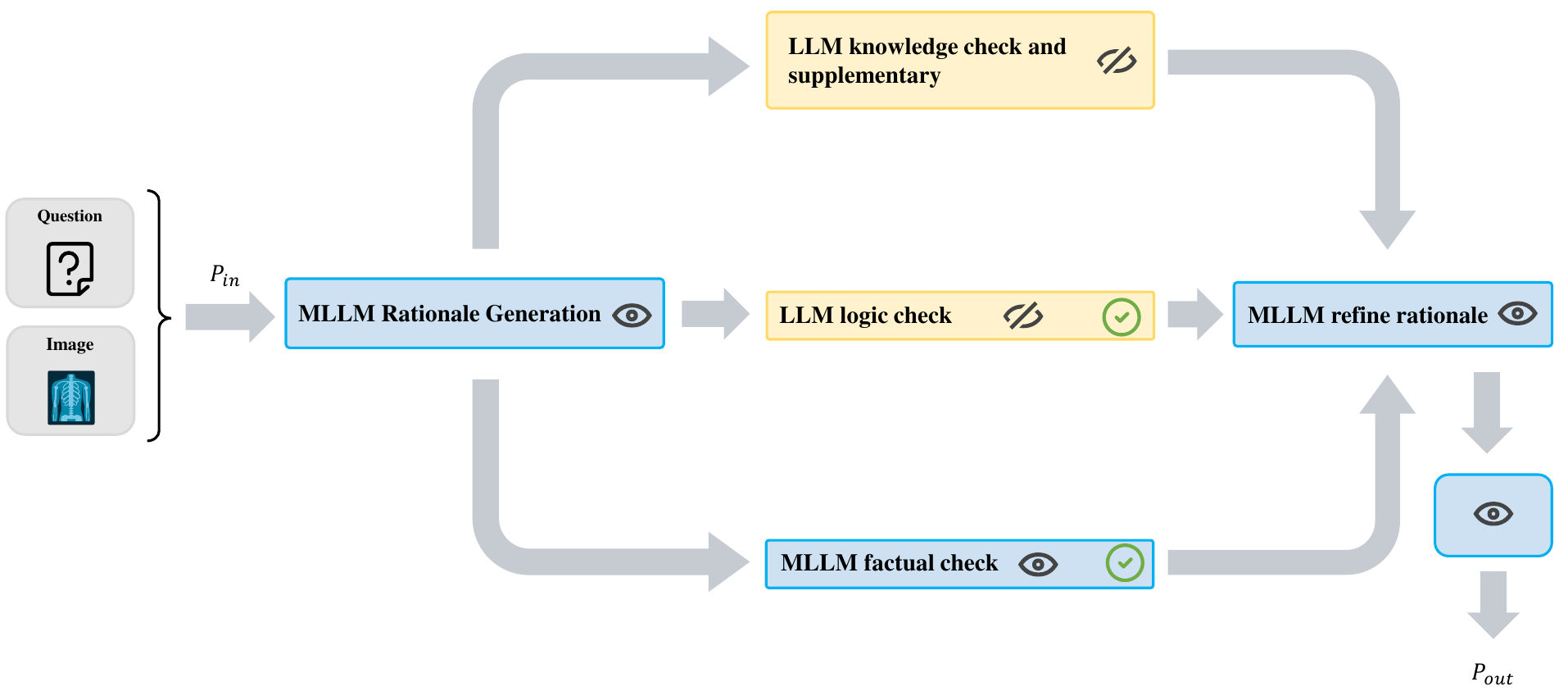}
    \caption{The schematic diagram of the FCCoT framework.}
    \label{fig:llm-check}
\end{figure*}

\textbf{FCCoT}\quad This framework employs both an LLM and an MLLM to check for flaws in the MLLM's output. The structure is illustrated in Figure \ref{fig:llm-check}. Initially, the MLLM provides an answer and a reasoning process based on the question. However, due to the limitations of the MLLM, this reasoning process may contain flaws. Potential flaws include factual inaccuracies, reasoning errors, and errors caused by a lack of background knowledge. We introduce an additional MLLM process to correct factual errors by verifying that the original reasoning aligns with the entities depicted in the image. The LLM is used to correct reasoning flaws and highlight any missing background knowledge. Finally, based on all the supplementary information, the MLLM regenerates the reasoning process and the answer.

\section{Prompts Used in MC-CoT}
\label{appendix:prompts}
\begin{figure*}
    \centering
    \includegraphics[width=1\linewidth]{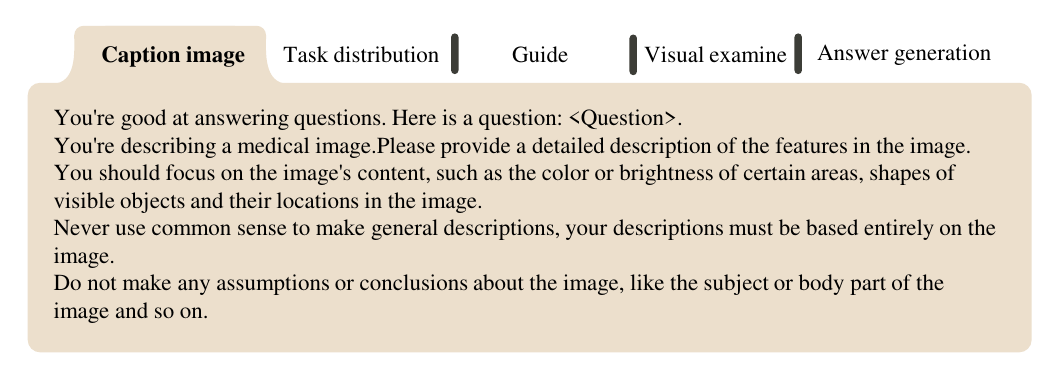}
    \caption{The prompt used to instruct MLLM generate a caption about the image.}
    \label{fig:prompt_cap}
\end{figure*}

\begin{figure*}
    \centering
    \includegraphics[width=1\linewidth]{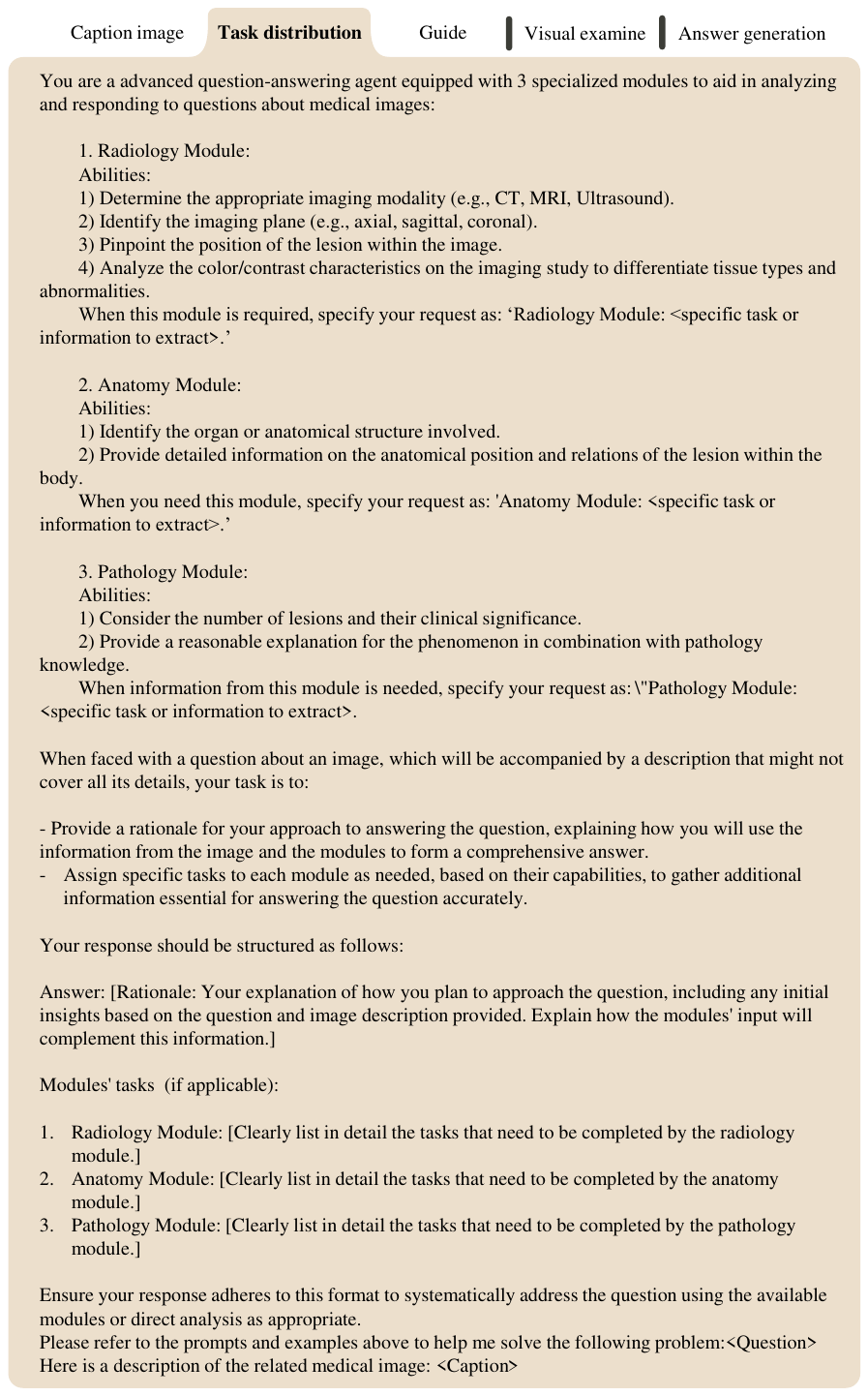}
    \caption{The prompt used to instruct LLM to distribute task to 3 carefully designed modules.}
    \label{fig:prompt_dis}
\end{figure*}

\begin{figure*}
    \centering
    \includegraphics[width=1\linewidth]{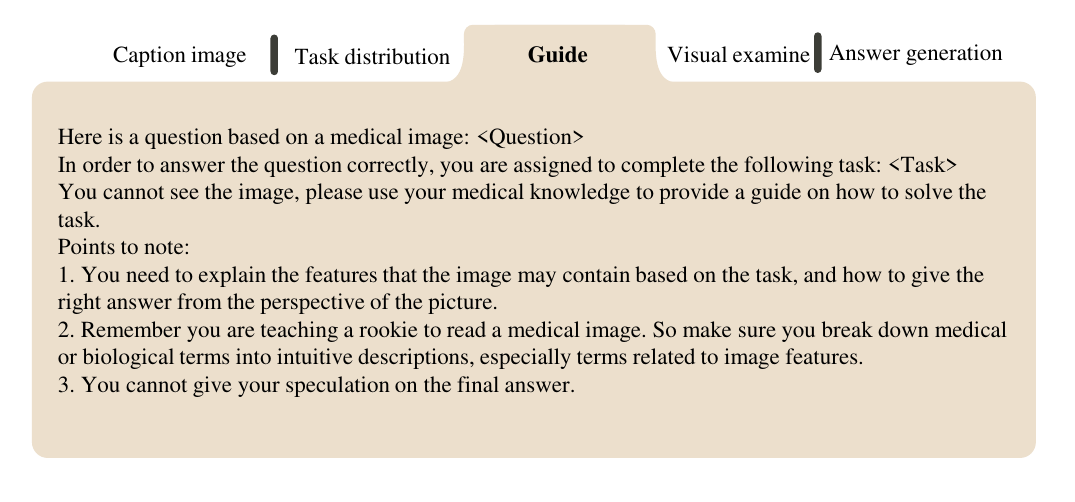}
    \caption{The prompt used to instruct LLM to give necessary background infomation as well as a guide on solving the problem.}
    \label{fig:prompt_guide}
\end{figure*}

\begin{figure*}
    \centering
    \includegraphics[width=1\linewidth]{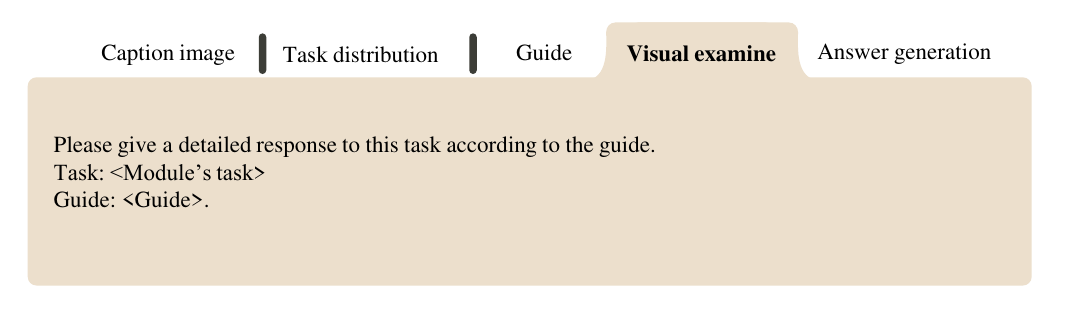}
    \caption{The prompt used to instruct MLLM to follow the guide and finish the assigned task.}
    \label{fig:prompt_v}
\end{figure*}

\begin{figure*}
    \centering
    \includegraphics[width=1\linewidth]{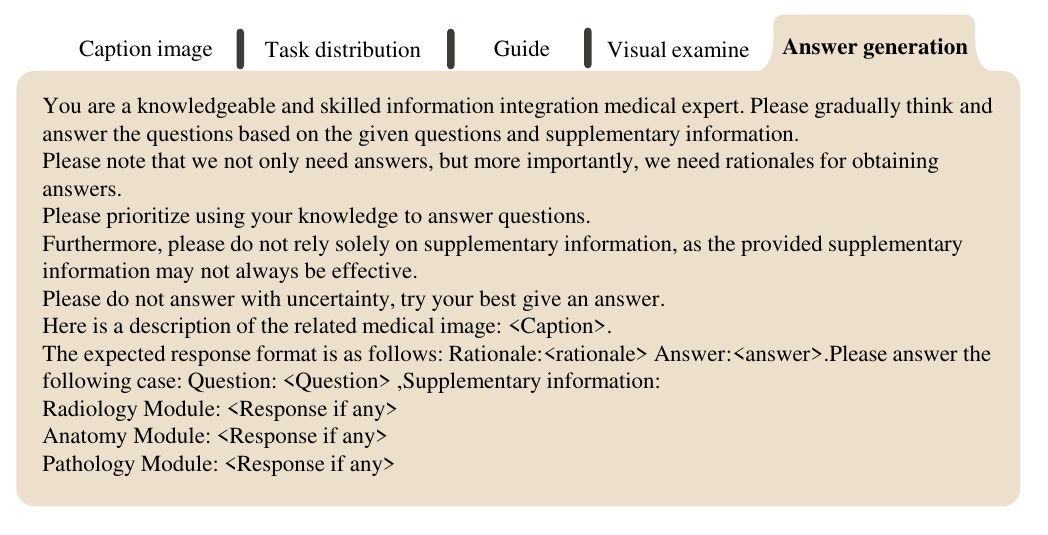}
    \caption{The prompt used to instruct LLM to generate the final answer.}
    \label{fig:prompt_answer}
\end{figure*}

The effectiveness of the MC-CoT framework largely depends on carefully designed prompts that guide the model through various stages. 
Below are examples of the prompts used in the MC-CoT framework:
\begin{itemize}
    \item Figure \ref{fig:prompt_cap} shows the prompt used to guide the MLLM in generating image descriptions.
    \item Figure \ref{fig:prompt_dis} shows the prompt used to guide the LLM in assigning questions to three specially designed modules.
    \item Figure \ref{fig:prompt_guide} shows the prompt used to guide the LLM in providing necessary background information and problem-solving guidance.
    \item Figure \ref{fig:prompt_v} shows the prompt used to guide the MLLM in following instructions and completing tasks.
    \item Figure \ref{fig:prompt_answer} shows the prompt used to guide the LLM in generating the final answer.
\end{itemize}
These prompts are designed to ensure that the model can accurately understand the problem, effectively extract information from the image, and generate precise and comprehensive answers.

\section{Evaluation Details}
\label{appendix:prompts_eval}
\begin{figure*}[t]
    \centering
    \includegraphics[width=1\linewidth]{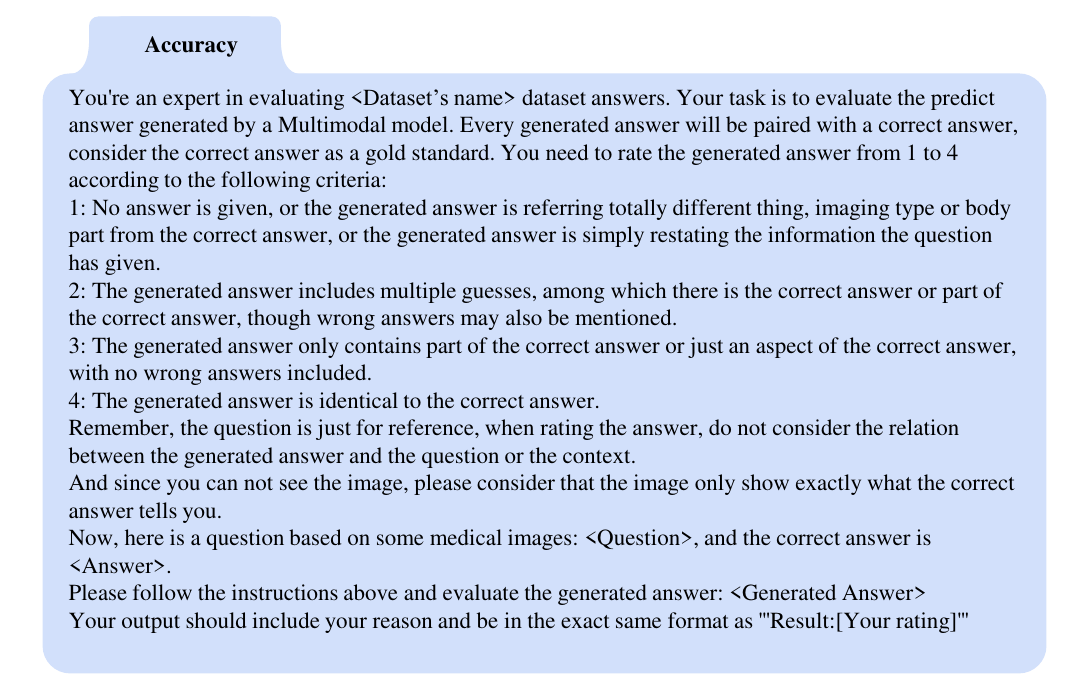}
    \caption{The prompt used to instruct Deepseek-V2 to evaluate the answers according to the given criteria.}
    \label{fig:prompt_eval}
\end{figure*}
We use Deepseek-V2 to assess the \emph{accuracy} of the model-generated answers. 

Given that the same medical concept can be expressed in various ways, we have developed a scoring system based on the degree of conceptual overlap. 

In this system, 1 point represents a refusal to answer or a completely incorrect answer, while 4 points indicate complete conceptual accuracy. 
Since we found that the model often provides multiple guesses when uncertain, which is undesirable, we assign 2 points to answers with multiple guesses and 3 points to answers with minor conceptual deviations.

We provide the correct answer along with the model-generated answer to Deepseek-V2 and use the prompt shown in Figure~\ref{appendix:prompts_eval} to instruct it to evaluate the model-generated answers.
Finally, the scores were scaled to 100 points.

On the other hand, for \emph{recall} calculation, since both MC-CoT and DDCoT include a rationale in their final outputs, which significantly increases the length of the answers, it would be unfair to calculate \emph{recall} based on this. 
Therefore, when calculating \emph{recall}, we excluded the rationale of MC-CoT and DDCoT and used only their final answers for the calculation.
\section{Example Cases}
\label{appendix:cases}

Through specific case analyses, we can more intuitively understand the advantages of the MC-CoT framework in handling open-ended Med-VQA tasks. 

\begin{figure*}
    \centering
    \includegraphics[width=1\linewidth]{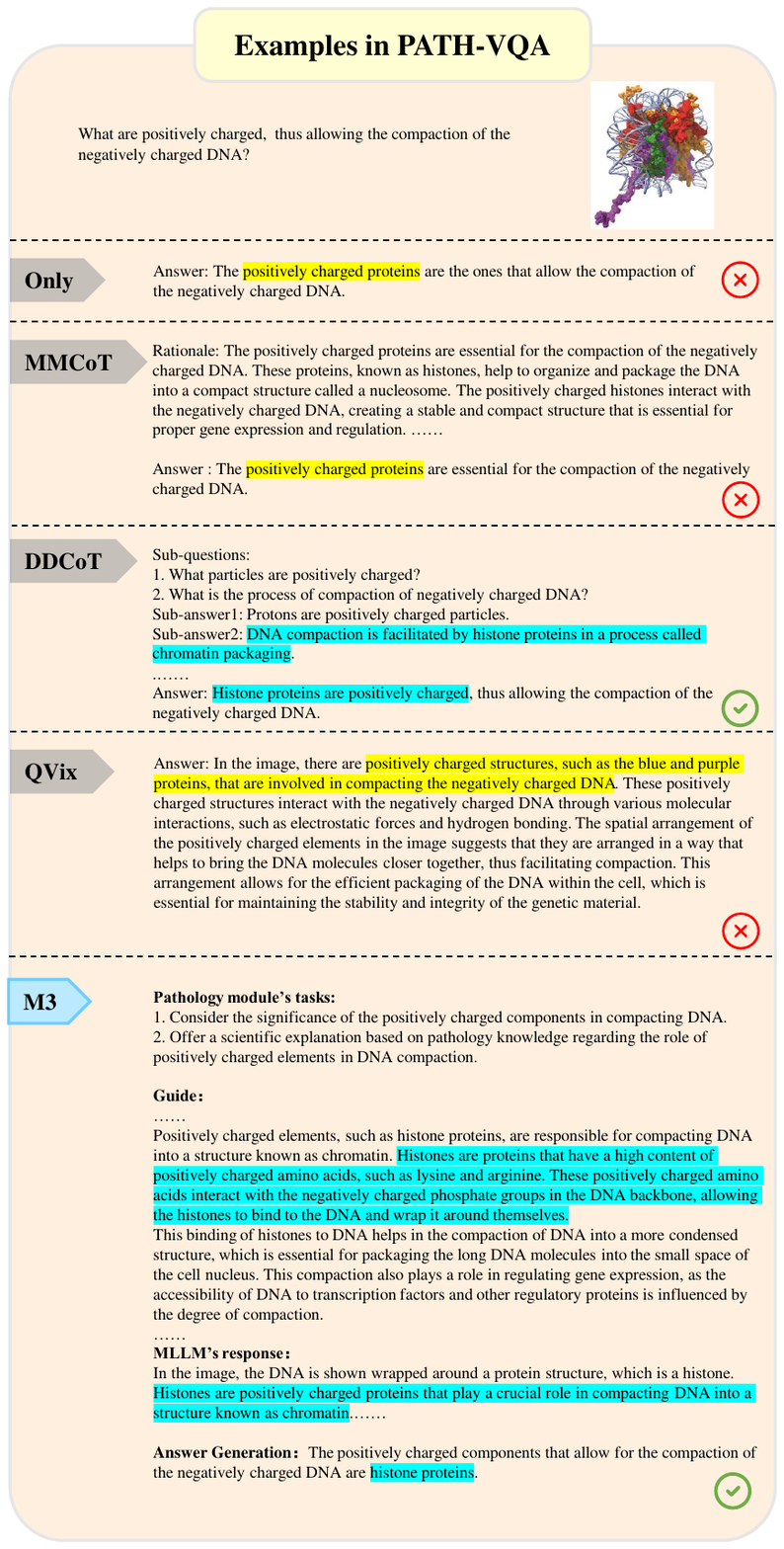}
    \caption{A specific example of the PATH-VQA dataset.}
    \label{fig:case_path}
\end{figure*}

Figure \ref{fig:case_path} shows a case from the PATH-VQA dataset, where the MC-CoT framework successfully generated an accurate answer through the reasoning process of the LLM. 

\begin{figure*}
    \centering
    \includegraphics[width=1\linewidth]{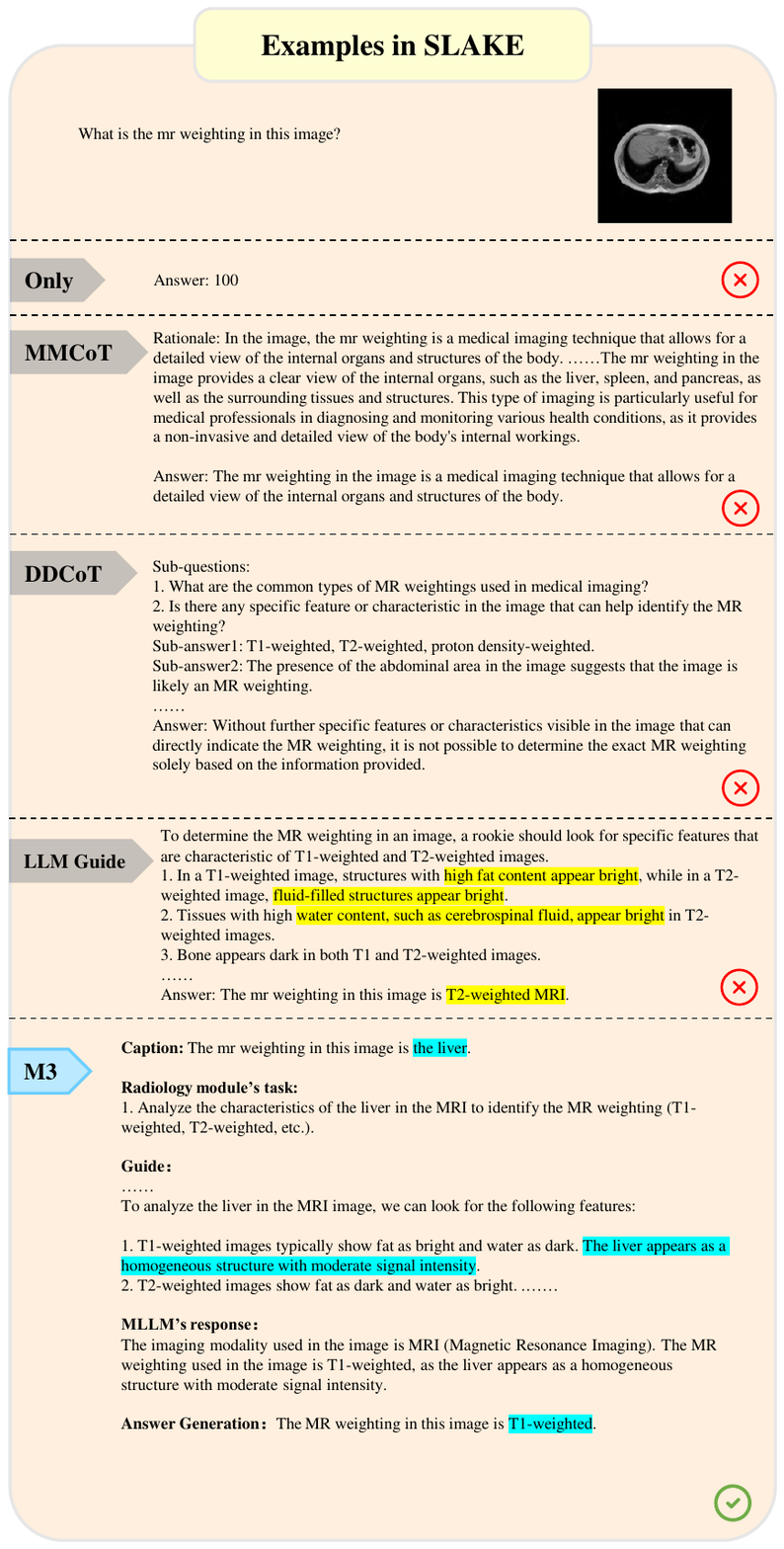}
    \caption{A specific example of the SLAKE dataset.}
    \label{fig:case_slake}
\end{figure*}

Figure \ref{fig:case_slake} shows a case from the SLAKE dataset, where the MC-CoT framework helped the LLM better understand the context of the question by generating an image description and providing targeted guidance. 

\begin{figure*}
    \centering
    \includegraphics[width=1\linewidth]{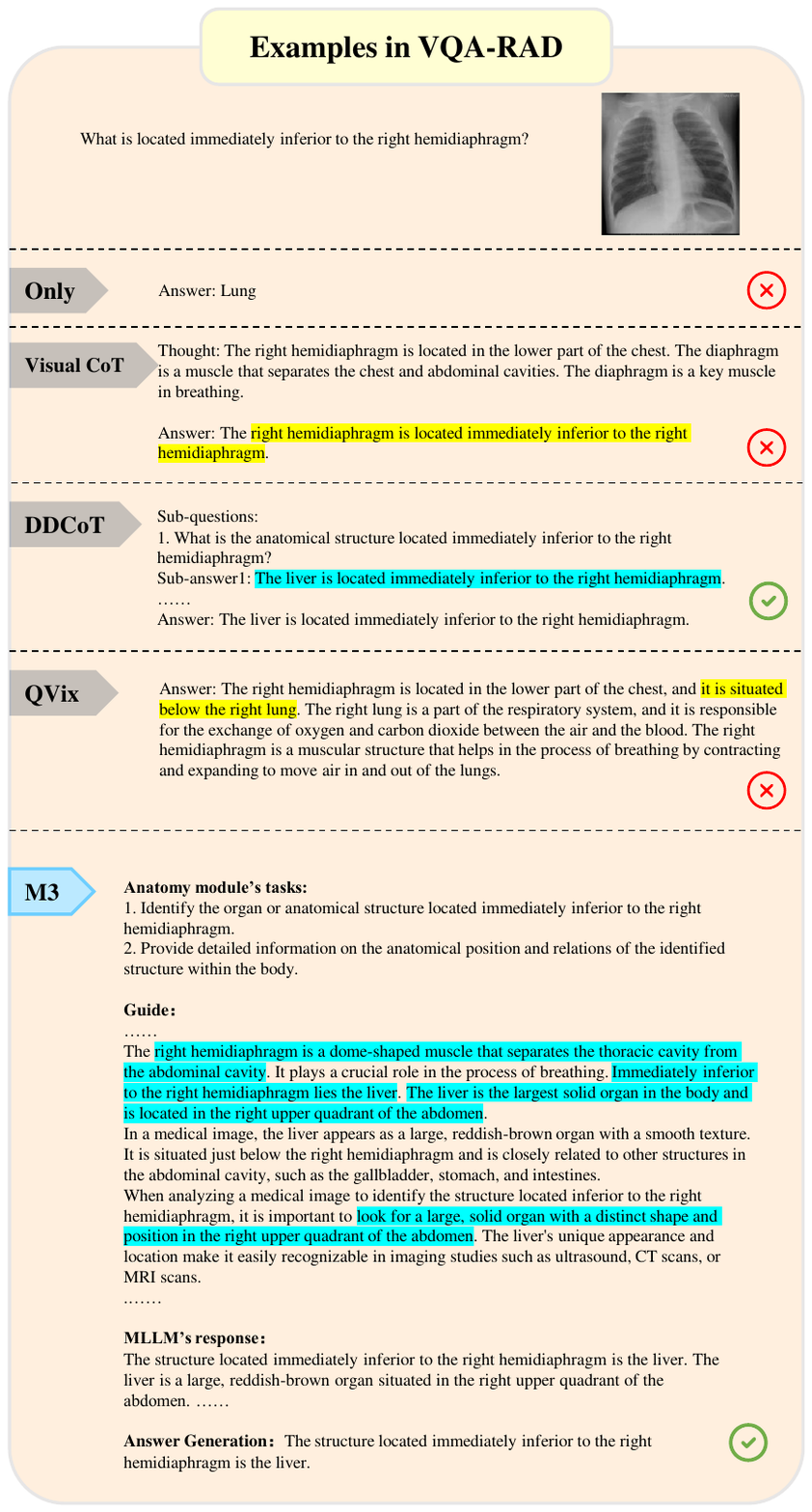}
    \caption{A specific example of the VQA-RAD dataset.}
    \label{fig:case_rad}
\end{figure*}

Figure \ref{fig:case_rad} shows a case from the VQA-RAD dataset, where the MC-CoT framework, with the guidance provided by the LLM, was able to accurately identify and interpret key features in the pathological image. 

These cases demonstrate the effectiveness and potential of the MC-CoT framework in handling complex and open-ended medical visual question-answering tasks.

%% file: Tables/change_mllm.tex
\begin{table*}
\centering
\begin{tabular}{l|cccccc|cc}
\toprule
\multicolumn{1}{l|}{}& \multicolumn{2}{c}{PATH-VQA} & \multicolumn{2}{c}{VQA-RAD} &\multicolumn{2}{c|}{SLAKE} &\multicolumn{2}{c}{Avg.}\\
& Recall & Acc.& Recall & Acc.& Recall & Acc.& Recall&Acc.\\
\midrule
\multicolumn{9}{c}{Qwen-VL-Chat with GPT-3.5} \\ 
\midrule
Only &43.92 & 25.87 & 53.60& 31.65 &68.91 & 51.47 &55.48 &36.33 \\
Visual CoT &44.90 & 19.95 &51.88 & 22.48 &69.56 & 44.34 &55.45 &28.92  \\
DDCoT &46.66 &34.08 &56.15 & \textbf{32.35} &\textbf{70.28} &48.63 &57.70  &38.35 \\
\textbf{MC-CoT\ (Ours)} &\textbf{50.02} & \textbf{35.52} & \textbf{57.72}&32.17 &70.19 &\textbf{51.73} &\textbf{59.31} &\textbf{39.81} \\ 
\midrule
\multicolumn{9}{c}{Deepseek-VL-7B with GPT-3.5}\\ 
\midrule
Only &42.55 & 25.49 & 51.96& 29.29 &68.65 &\textbf{54.57} &54.39 &36.45 \\
Visual CoT & 47.52& 31.15 &55.84 & \textbf{33.61} &72.54 & 51.68 &58.63 &38.81 \\
DDCoT&47.26 & 31.95 & 55.99& 29.22 & 70.57&45.94 &57.94 &35.70 \\
\textbf{MC-CoT\ (Ours)} &\textbf{49.44} & \textbf{35.47} &\textbf{58.83} & 31.96&\textbf{72.81} &52.71 &\textbf{60.36} &\textbf{40.05} \\ 
\midrule
\multicolumn{9}{c}{Qwen-VL-Max with GPT-3.5}\\ 
\midrule
Only &46.15 & 30.56 &58.82 &\textbf{38.88} & \textbf{73.40}& \textbf{59.33} &59.46 &42.92 \\
Visual CoT & 43.20&21.87 & 52.94&30.28 & 67.77& 52.04 &54.64 &34.73 \\
DDCoT &47.60 &32.32 &57.49 & 34.46& 70.75& 50.59 &58.61  &39.12 \\
\textbf{MC-CoT\ (Ours)}& \textbf{51.05}&\textbf{38.67} &\textbf{60.16} & \textbf{38.88} &71.35 & 54.99 &\textbf{60.85} &\textbf{44.18}\\ 
\bottomrule
\end{tabular}
\caption{Using different MLLMs to validate the effectiveness of the MC-CoT framework.}
\label{tab:change-mllm}
\end{table*}

%% file: Tables/change_llm.tex
\begin{table*}
\centering
\begin{tabular}{l|cccccc|cc}
\toprule
\multicolumn{1}{l|}{}& \multicolumn{2}{c}{PATH-VQA} & \multicolumn{2}{c}{VQA-RAD} &\multicolumn{2}{c|}{SLAKE} &\multicolumn{2}{c}{Avg.}\\
& Recall & Acc.& Recall & Acc.& Recall & Acc.& Recall&Acc.\\
\midrule
\multicolumn{9}{c}{LLaVA-v1.5-7B with GLM-4-9B-Chat} \\ 
\midrule
QVix & 49.37&34.88& 54.49&32.07 &69.20 & 48.06 & 57.69& 38.34\\
DDCoT & 49.47& 22.13 &54.16 & 24.31 &67.34 & 36.43 &56.99 & 27.62\\
IICoT &49.51 & 40.00 &55.64 & 33.83 &68.62 & 43.26 &57.92 & 39.03\\
\textbf{MC-CoT\ (Ours)} & \textbf{52.40}& \textbf{42.08} &\textbf{58.55} & \textbf{40.96} & \textbf{71.20}& \textbf{54.68} & \textbf{60.72}& \textbf{45.91}\\ 
\midrule
\multicolumn{9}{c}{LLaVA-v1.5-7B with Qwen2-72B-Instruct}\\ 
\midrule
QVix &\textbf{49.29} & 31.31 & 53.40& 31.61 & \textbf{69.36}& 48.79 &57.35 & 37.24\\
DDCoT &44.79 & 29.87 &50.88 & 28.94 &65.95 & 44.65 &53.87 &34.49 \\
IICoT &50.04 & \textbf{36.91} & \textbf{55.42}& 32.81 & 68.38& 49.10 &\textbf{57.95} &39.61 \\
\textbf{MC-CoT\ (Ours)} & 48.97& 36.21 & 54.97& \textbf{38.32} & 69.08& \textbf{54.50} & 57.67& \textbf{43.01}\\  
\midrule
\multicolumn{9}{c}{LLaVA-v1.5-7B with Deepseek-V2}\\ 
\midrule
QVix &49.13 & 33.71 &53.44 & 31.89 & 69.01& 49.04 & 57.19& 38.21\\
DDCoT & 39.45& 30.77 &45.99 & 28.21 & 60.35& 37.57 &48.59 & 32.18\\
IICoT & 47.91& 36.91 &55.40 & 33.09 & 69.37& 46.51 & 57.56& 38.84\\
\textbf{MC-CoT\ (Ours)} &\textbf{49.87} & \textbf{49.17} & \textbf{56.52}& \textbf{48.93} & \textbf{70.71}& \textbf{60.31} &\textbf{59.03} &\textbf{52.80} \\ 
\bottomrule
\end{tabular}
\caption{Using different LLMs to validate the effectiveness of the MC-CoT framework.}
\label{tab:change-llm}
\end{table*}